\documentclass[11pt]{article}

\usepackage[margin=1in]{geometry}
\usepackage[T1]{fontenc}
\usepackage[utf8]{inputenc}
\usepackage{graphicx}
\usepackage{booktabs}
\usepackage{multirow}
\usepackage{subcaption}          
\usepackage{amsmath,amssymb}
\usepackage{xcolor}
\usepackage{hyperref} 
\usepackage{cleveref}            
\usepackage{orcidlink}
\usepackage{url}

\begin{document}

\title{Mitigating Domain Shift in Conditioned Floor Plan Generation:\\
Synthetic Pre-training for Data-Efficient Adaptation}

\author{
Matthieu Ospici$^{1}$ \and
Arnaud Gueze$^{1,2}$ \and
Luc Bourrat$^{1}$ \and
Adrien Bernhardt$^{1}$ \\[4pt]
{\normalsize $^{1}$Homiwoo, 8 rue la Bo\'etie, 75008 Paris, France} \\
{\normalsize $^{2}$\'Ecole Polytechnique, Route de Saclay, 91128 Palaiseau, France} \\
{\small \url{https://www.homiwoo.com/}}
}
\date{}

\maketitle

\begin{abstract}

Robustness to domain shift is a key requirement for floor plan generative models to be applicable beyond the single dataset they were trained on, as floor plans vary widely across regions due to distinct architectural cultures, spatial constraints, and construction practices, while acquiring new annotated datasets remains costly and domain-specific.  Yet, no prior work has studied this robustness in the context of conditioned floor plan generation. In this paper, we evaluate state-of-the-art models from two fundamentally different generative paradigms across three public datasets (RPLAN, MagicPlan and Swiss Dwellings) and show that they are highly sensitive to domain shift, with up to an order of magnitude performance degradation when transferred across domains.
To mitigate this with minimal target-domain supervision, we introduce a procedural method to generate a large-scale synthetic training dataset that enforces strict physical constraints (non-overlapping rooms, valid door placement, graph consistency) while intentionally sacrificing architectural realism through highly irregular spatial arrangements and aggressive geometric perturbation of room shapes. We show that pre-training on this synthetic data considerably improves zero-shot cross-domain performance, outperforming in-domain training on MagicPlan. Furthermore, it provides a highly effective initialization for fine-tuning, accelerating target domain adaptation and outperforming real-world initialization baselines by up to 40\% in a low-data regime.
\end{abstract}

\section*{Introduction}
\label{sec:intro}

Floor plan generation conditioned on architectural constraints, such as room shapes and user-specified connectivity graphs, has recently attracted considerable interest~\cite{puzzlefusion, housediffusion, nauata2020house, dpfm, gueze2023floor}.
These advances have, for the most part, been sparked by the rise of Diffusion Models ~\cite{ho2020denoising, song2021scorebased} and more recently Flow Matching methods~\cite{lipmanflow, liu2023flow, albergo2023stochastic}, driving significant progress in this field. Existing approaches can be broadly categorized into two paradigms: arrangement-based methods\cite{dpfm, puzzlefusion}, which generate layouts by predicting rigid transformations of input polygons, and coordinate-based methods \cite{gueze2023floor, housediffusion, polydiffuse}, which directly diffuse polygon vertices under graph constraints. Despite their architectural differences, all existing methods are evaluated on a single dataset, leaving the question of cross-domain generalization unexplored.

Available floor plan datasets \cite{puzzlefusion, rplan, swiss, msd, cubicasa} reflect a distinct acquisition process and architectural culture, inducing different geometric distributions, room shape statistics, and topological conventions. Collecting new annotated floor plans is costly, as it requires professional drafting, mobile scanning, or manual digitization with quality assurance, making it impractical to build large-scale datasets for every target domain. We show through a systematic cross-domain evaluation that models trained on any single dataset consistently fail to generalize to the others, regardless of the direction of transfer: models overfit to the specific spatial regularities of their training domain rather than learning the fundamental rules of layout assembly. To the best of our knowledge, this domain shift problem in conditioned floor plan generation has never been formally studied, and cross-domain transferability has not been evaluated in this setting.

To test whether this failure mode is specific to one paradigm, we select two models representing the two dominant vector-based generative paradigms. The first is an arrangement-based Flow Matching model~\cite{dpfm}, which treats layout generation as a set of rigid transformations predicting the translation and rotation of input polygons. The second is a constraint-based Diffusion model~\cite{gueze2023floor}, which operates at the vertex level, diffusing all polygon points simultaneously while reconciling local shapes with global graph constraints via cross-attention. Their shared vulnerability to domain shift indicates that the problem is consistent across the two paradigms we study rather than tied to a specific architectural choice.

To bridge this gap, we propose a novel pre-training strategy based on large-scale, procedurally generated synthetic layouts. Our key insight is that generalization requires decoupling the learning of spatial assembly rules from statistical regularities of any particular architectural style. 
To this end, our synthetic pipeline deliberately abandons architectural plausibility. We extract basic room shapes and subject them to geometric perturbations: arbitrary bumps, hollows, and heavy scale variations, before assembling them densely, with doors assigned at shared wall segments. 
The resulting configurations bear no visual resemblance to realistic floor plans, yet they enforce the underlying physical constraints of layout generation: non-overlapping geometries, valid door placements, and exact adherence to the connectivity graph. Pre-training on this implausible dataset prevents models from exploiting dataset-specific shortcuts and forces them to internalize the combinatorial logic of spatial assembly. We show that this strategy significantly improves both fine-tuning performance on scarce real-world data and zero-shot transfer across domains, for both modeling paradigms.
In summary, our main contributions are as follows:

\begin{itemize}
  \item \textbf{Cross-Domain Study:} We formalize the cross-domain transfer problem in conditioned floor plan generation and present a cross-dataset study on RPLAN~\cite{rplan}, MagicPlan~\cite{puzzlefusion} and Swiss Dwellings~\cite{swiss}, evaluated with geometric and topological metrics (MPE, NGED).
\item \textbf{Consistent Failure Across Paradigms:} We show that performance collapse under domain shift is consistent across two fundamentally different generative paradigms, a room-wise rigid-based model~\cite{dpfm} and a vertex-wise constraint-based model~\cite{gueze2023floor}.
\item \textbf{Synthetic Pre-training for Data-Efficient Adaptation:} We propose a procedural data generation pipeline that intentionally sacrifices architectural realism in favor of geometric and topological diversity, while maintaining physical constraints. We show that pre-training on these implausible layouts enables competitive zero-shot transfer, even surpassing complete in-domain training on MagicPlan with DPFM. Furthermore, it provides a highly effective initialization for fine-tuning, accelerating target domain adaptation.
\end{itemize}

\section{Related Work}
\paragraph{Floor Plan Generation.} Generating floor plans as sets of constrained polygons has been approached through a variety of deep generative frameworks.  Early approaches operated on raster representations, including HouseGAN, HouseGAN++~\cite{nauata2020house,nauata2021house} and Graph2Plan~\cite{graph2plan}. More recently, vectorial approaches have been proposed:  HouseDiffusion \cite{housediffusion} introduced a diffusion model \cite{ho2020denoising, song2021scorebased} operating at the vertex level, conditioning generation on a bubble diagram. PolyDiffuse \cite{polydiffuse} addressed per-polygon noise heterogeneity through dedicated guidance networks. Gueze et al.~\cite{gueze2023floor} further constrained the diffusion process using potentially partial room shapes as input, incorporating cross-attention mechanisms to enforce the constraints during the generation. In parallel, arrangement-based approaches reframed the problem as spatial assembly: rather than generating coordinates from noise, they predict rigid transformations to position predefined room shapes into a coherent layout \cite{puzzlefusion}. More recently, DPFM~\cite{dpfm} proposed a unified framework that combines rigid transformations with vertex-level operations. Nevertheless, this rapid progress has been exclusively measured within single datasets, leaving the issue of cross-domain robustness completely unaddressed.

\paragraph{Domain Shift and Generalization.} Domain shift has been extensively studied in recognition tasks, where unsupervised domain adaptation~\cite{coral,mmd,reversal} and domain generalization~\cite{zhou2022} have demonstrated strong results. However, classical unsupervised adaptation is inapplicable to conditioned layout generation, as the target floor plan cannot be disentangled from its conditioning inputs (the graph and polygons), leaving no usable unannotated signal. Similarly, domain generalization spans both multi-source and single-source settings~\cite{zhou2022}. Multi-source DG assumes several annotated domains, impractical here given the scarcity of public floor plan datasets~\cite{cubicasa, msd, puzzlefusion, rplan, lifull}. Single-source DG typically augments a single real domain to broaden coverage of the target distribution; our setting differs in that the synthetic prior does not aim to cover or resemble any target distribution, and transfer instead relies on structural constraints shared across domains.

In the sim-to-real setting, domain randomization~\cite{dom_rand, dom_rand_tremblay} achieves domain generalization using heavily randomized environments bearing little visual resemblance to the target domain. Adapting this principle, we propose a novel synthetic pre-training strategy to mitigate domain shift in conditioned floor plan generation.

\paragraph{Synthetic Data for Generalization.} Procedurally generated synthetic data has proven effective as a pre-training source in settings where real annotated data is scarce or domain-specific. In computer vision, synthetic rendering has been used to bootstrap models for depth estimation \cite{Mayer_2016_CVPR}, object detection \cite{obj}, and scene understanding \cite{McCormac}.  In the specific context of floor plan generation, procedurally generated datasets do 
exist. ProcTHOR~\cite{procthor} or Aria \cite{aria_data_tools} generate large-scale synthetic indoor environments 
for embodied AI, and Structure3D~\cite{Structured3D} provides synthetically rendered 
floor plans with rich annotations. These approaches prioritize architectural 
plausibility: room shapes are simple and regular, and layouts are designed to closely 
mimic realistic floor plans. In contrast, our approach deliberately abandons architectural plausibility, maximizing geometric diversity to build a  style-agnostic prior rather than mimicking any particular target distribution.

\section{Method}
\subsection{Problem Formulation}
\label{sec:probform}
Let $\mathcal{D} = \{(G_i, L_i)\}$ denote a floor plan dataset, where 
$G_i = (\mathcal{V}_i, \mathcal{E}_i)$ is a connectivity graph whose nodes represent 
rooms and edges represent doorways between adjacent rooms, each associated with a pair 
of door segments lying on the shared wall boundary. $L_i = \{\mathcal{P}_i^k\}_{k=1}^{|\mathcal{V}_i|}$ denotes the layout composed of room polygons $\mathcal{P}_i^k$.
We train a generative model $f_\theta$ to estimate $L_i$ from $G_i$ and the set of unposed room polygons $P_i = \{P_i^k\}$ so that $L_i \sim f_\theta(\cdot \mid G_i, P_i)$.

The unposed rooms must be spatially arranged into a coherent floor plan while satisfying both 
the geometric constraints imposed by the room shapes and the topological constraints 
encoded in the graph. Examples of coherent floor plans can be seen in \Cref{fig:dataset_comparison}, where rooms are arranged without overlapping while strictly respecting the connectivity graph.

We consider a multi-domain setting with datasets $\{\mathcal{D}^d\}_{d=1}^D$, each 
reflecting a distinct acquisition process and architectural differences, inducing a 
domain-specific distribution $p^d(G, L)$ over graphs and layouts. We define 
the domain shift problem as follows: a model $f_\theta$ trained on source domain 
$\mathcal{D}^s$ is evaluated on a target domain $\mathcal{D}^t \neq \mathcal{D}^s$. 
Domain shift occurs when:
\begin{equation}
    \mathcal{L}(f_\theta, \mathcal{D}^t) \gg \mathcal{L}(f_\theta, \mathcal{D}^s)
\end{equation}
where $\mathcal{L}$ denotes a layout quality metric (MPE or NGED). We show in 
Section~\ref{sec:experiments} that this inequality holds for all pairs $(s, t)$ across the three datasets, regardless of the direction of transfer, establishing that domain shift in floor plan generation is a systematic and bidirectional phenomenon.

To address this, we propose to pre-train $f_\theta$ on a large synthetic dataset 
$\mathcal{D}^{\text{synt}}$, designed to minimize dependence on any single domain, before fine-tuning on any 
target domain $\mathcal{D}^t$. The key property of $\mathcal{D}^{\text{synt}}$ is that 
it enforces: 
non-overlapping geometries, valid door placements, and graph consistency,  while 
deliberately avoiding any domain-specific geometric regularities.

\subsection{Synthetic Pre-training Dataset}
\label{sec:pre_train_dataset}
\begin{figure}[t]
\includegraphics[width=\textwidth]{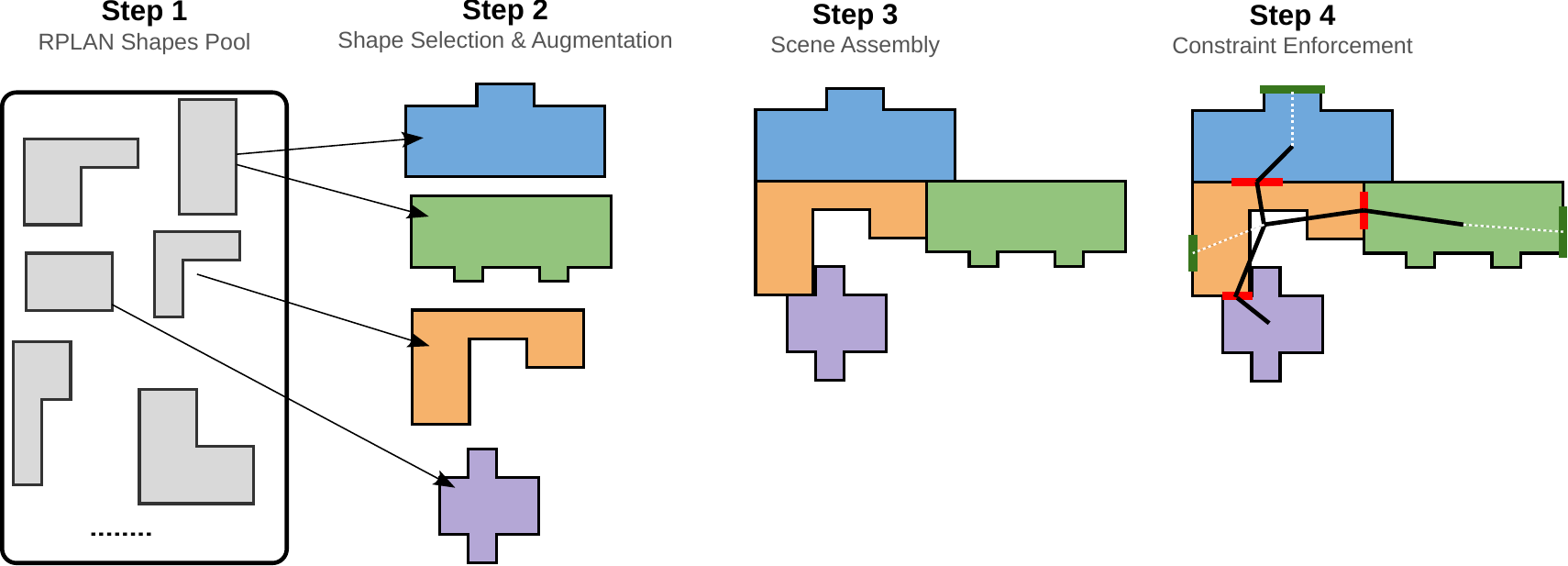}
\centering
\small
\caption{Procedural generation of the synthetic pre-training dataset. Our pipeline deliberately abandons architectural realism to construct a geometric environment that forces models to learn the fundamental rules of spatial assembly. Isolated room polygons are first sampled from a base dataset to serve as shape priors. These shapes undergo geometric augmentations, introducing arbitrary concavities, convexities, and scale distortions, before being densely packed to maximize shared boundaries without overlaps. Finally, interior doors (red) and exterior windows (green) are instantiated on valid wall segments, and the corresponding topological graph is created to condition the generative models.}
\label{fig:synth}
\end{figure}

\paragraph{Overview.} Our synthetic dataset is generated procedurally from individual room shapes sampled from RPLAN. Crucially, only isolated room polygons are extracted, no connectivity, spatial arrangement or doors/windows position is retained. Rather than generating shapes from scratch, which risks producing geometrically degenerate polygons with no relation to plausible room geometry, RPLAN serves as a weak prior on basic shape properties such as rectilinearity, orthogonality and parallelism. 
The subsequent augmentation pipeline is designed to break any remaining statistical link with the RPLAN distribution. The resulting scenes no longer resemble their source in terms of shape statistics, scale, or spatial arrangement. The synthetic layouts are geometrically implausible as floor plans, yet strictly enforce the physical constraints of spatial assembly: non-overlapping rooms, valid door placements on shared walls, and exact adherence to a connectivity graph.
\paragraph{Shape augmentation.} Each sampled polygon is independently perturbed through four successive transformations. First, an aspect ratio distortion stretches the shape along its principal axis by a random factor in [0.8, 1.25]. Second, one to three rectangular bumps (protrusions or indentations) are inserted on randomly selected edges, introducing concavities and convexities absent from the original shapes. Third, random flips and rotations in $\{0^\circ, 90^\circ, 180^\circ, 270^\circ\}$
 break any canonical orientation bias. Finally, a global scale factor sampled uniformly in 
[0.7, 1.6] introduces significant size variance across rooms. All transformations preserve rectilinearity and polygon validity.

\paragraph{Scene assembly.} Augmented rooms, between 2 and 10 per scene, sampled according to a distribution weighted toward medium-density layouts, are placed sequentially using an edge-packing strategy that maximizes shared boundary length while strictly prohibiting overlaps. Rooms are placed with a random offset from the centroid of existing rooms, preventing the repetitive cross-shaped patterns that emerge when all rooms are aligned around a central piece. After placement, a separation pass eliminates any residual overlaps.

\paragraph{Constraint enforcement.} Interior doors are placed randomly on all shared wall segments identified after assembly, using a mixture strategy that samples door lengths as a fraction of the shared boundary (small: 
[0.18, 0.35], medium: 
[0.35, 0.55], large: 
[0.55, 0.85]). Door endpoints are snapped to room boundaries with a tolerance of 
0.03 in the normalized [-1,1] coordinate system. Each doorway pair is validated for midpoint proximity and parallelism before the scene is accepted. Scenes where any door pair cannot be reconciled are rejected. A small asymmetry is additionally introduced between the two segments of each doorway pair to simulate the measurement noise present in real datasets.

\paragraph{Scale and diversity.} The synthetic dataset comprises 135k scenes.  Figure~\ref{fig:synth} summarizes the pipeline implemented for the creation of our synthetic dataset. Despite sharing prior shapes with RPLAN, the resulting layouts occupy a fundamentally different geometric distribution, as shown in \Cref{tab:dataset_stats} and visually in \Cref{fig:dataset_comparison}.

\section{Experiments}
\label{sec:experiments}

\subsection{Experimental Setup}

\paragraph{Datasets} We evaluate our approach on three real-world datasets covering distinct architectural cultures and acquisition processes, alongside our proposed Synthetic dataset. 

\begin{figure}[t]
\includegraphics[width=\textwidth]{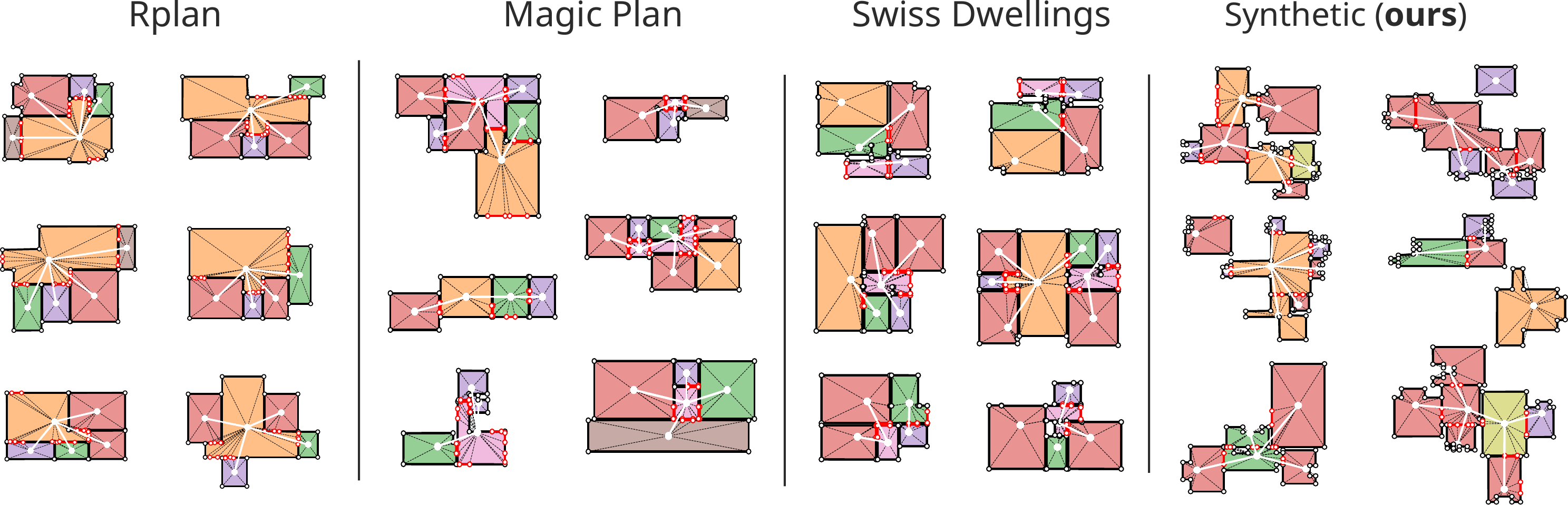}
\centering
\small
\caption{Qualitative comparison of the evaluated domains. Real-world datasets (Rplan,  Magic Plan, Swiss Dwellings) exhibit regular rectangular packings characteristic of their respective acquisition processes. Our Synthetic dataset (right) introduces extreme geometric and layout variance. Quantitative differences are detailed in \Cref{tab:dataset_stats}.}
\label{fig:dataset_comparison}
\end{figure}

\begin{description}
    \item[RPLAN~\cite{rplan}] contains approximately 60k annotated floor plans of Asian residential apartments, derived from professional architectural drawings, characterized by compact room arrangements and strong geometric regularity. 
    \item[MagicPlan~\cite{puzzlefusion}] originally comprises 90k floor plans collected via a mobile indoor scanning application. We preprocess it into a Manhattan geometry representation to ensure compatibility across both evaluated models, resulting in approximately 40k valid floor plans. 
    \item[Swiss Dwellings~\cite{swiss}] contains Swiss residential floor plans digitized from existing building plans by a commercial provider (Archilyse AG), with manual quality assurance ensuring geometric accuracy. Following a similar spirit to MSD~\cite{msd}, we filter multi-floor buildings, perform deduplication, retain Manhattan samples, and extract missing door segments. This yields a processed split of approximately 5k floor plans. Due to its limited size, Swiss Dwellings is used exclusively for target evaluation and fine-tuning. 
\end{description}

\Cref{tab:dataset_stats} details the topological and geometric statistics of these four datasets, with qualitative samples provided in \Cref{fig:dataset_comparison}.
As illustrated in the Figure, the extracted topological graphs accurately reflect the inherent noise of real-world dataset acquisition. Rather than being connected networks, the graphs can contain disconnected components and isolated rooms, a phenomenon explicitly quantified by the \textit{Connected components} and \textit{Isolated rooms} metrics in \Cref{tab:dataset_stats}. This topology primarily arises from measurement errors in door localization.

\begin{table}[t]
    \centering
    \small
    \caption{Comparison of dataset statistics (mean values per floor plan). Our synthetic dataset exhibits significantly higher geometric complexity at the room level (8.83 corners per room) compared to real-world datasets, reflecting the aggressive augmentation pipeline described in \Cref{sec:pre_train_dataset}.}
    \label{tab:dataset_stats}
    \begin{tabular}{lcccc}
        \toprule
        \textbf{Metric (Mean)} & \textbf{RPLAN} & \textbf{MagicPlan} & \textbf{Swiss Dwellings} & \textbf{Synthetic} \\
        & \cite{rplan} & \cite{puzzlefusion} & \cite{swiss} & \textbf{(Ours)} \\
        \midrule
        Number of rooms & \textbf{6.84} & 5.63 & 6.20 & 4.42 \\
        Connected components & 3.54 & 4.42 & \textbf{4.88} & 3.71 \\
        Isolated rooms & 2.44 & 3.77 & \textbf{4.16} & 3.18 \\
        Corners per room & 5.32 & 4.67 & 6.97 & \textbf{8.83} \\
        \bottomrule
    \end{tabular}
\end{table}

\paragraph{Models} All experiments are conducted with two models representative of the 
two dominant vectorial generative paradigms: DPFM~\cite{dpfm}, an arrangement-based Flow Matching 
model that predicts rigid transformations of input polygons, and Gueze et al.~\cite{gueze2023floor}, 
a constraint-based Diffusion model that operates at the vertex level. Both models are conditioned on the connectivity graph and input room shapes 
as described in Section~\ref{sec:probform}. Rather than comparing absolute performance 
between the two models, which differ in architecture and training procedure, we focus on 
consistent trends across both paradigms.

\paragraph{Metrics} We evaluate generated layouts using two complementary metrics. Mean Position Error (MPE $\downarrow$) measures the average Euclidean distance between predicted and ground-truth room vertices in a $256 \times 256$ pixel space, following prior work~\cite{puzzlefusion, dpfm}. The metric is computed after optimal rigid alignment (translation and rotation), making it invariant to global pose ambiguity. Rather than the raw GED used in~\cite{puzzlefusion, dpfm}, we report a  Normalized Graph Edit Distance (NGED $\downarrow$), which captures topological correctness by comparing the predicted connectivity graph against the ground truth. We normalize by the edge counts, yielding values in $[0, 1]$ comparable across layouts of varying complexity. 

\paragraph{Evaluation protocol} All experiments are conducted under full graph conditioning. Furthermore, to simulate the measurement inaccuracies ubiquitous in real-world acquisitions, we inject structural noise into the door constraints, specifically, randomized door lengths, across all domains and all models. This unified  setup deviates from the original DPFM benchmark, which operates without graph conditioning and without door noise; our reproduced DPFM baseline achieves an 
intra-domain MPE of $6.31$ on RPLAN, compared to approximately $3.0$ in the original 
unconstrained setting. We maintain this rigorous protocol across all experiments to 
ensure a fair and consistent measurement of domain shift and transferability.

\paragraph{Pre-training Strategy}



Both evaluated models~\cite{gueze2023floor, dpfm} are pre-trained on $\mathcal{D}^{\text{synt}}$ following their standard procedures, without modifying the optimizer, learning rate schedule, or batch size. 

We evaluate under two regimes: \textit{zero-shot} (direct evaluation on target domain $\mathcal{D}^t$) and \textit{fine-tuning} (training on~$N$ target examples). For fine-tuning, the learning rate is reduced by a factor of 10. We fine-tune for a fixed 50k steps across all setups; empirically, we observed no overfitting, with prolonged training yielding stable performance. Models are trained on NVIDIA H100 GPUs.


\subsection{Zero-Shot Cross-Domain Transfer}
\label{sec:zero_shot}
\begin{figure}[t]
    \centering
    \includegraphics[width=\textwidth]{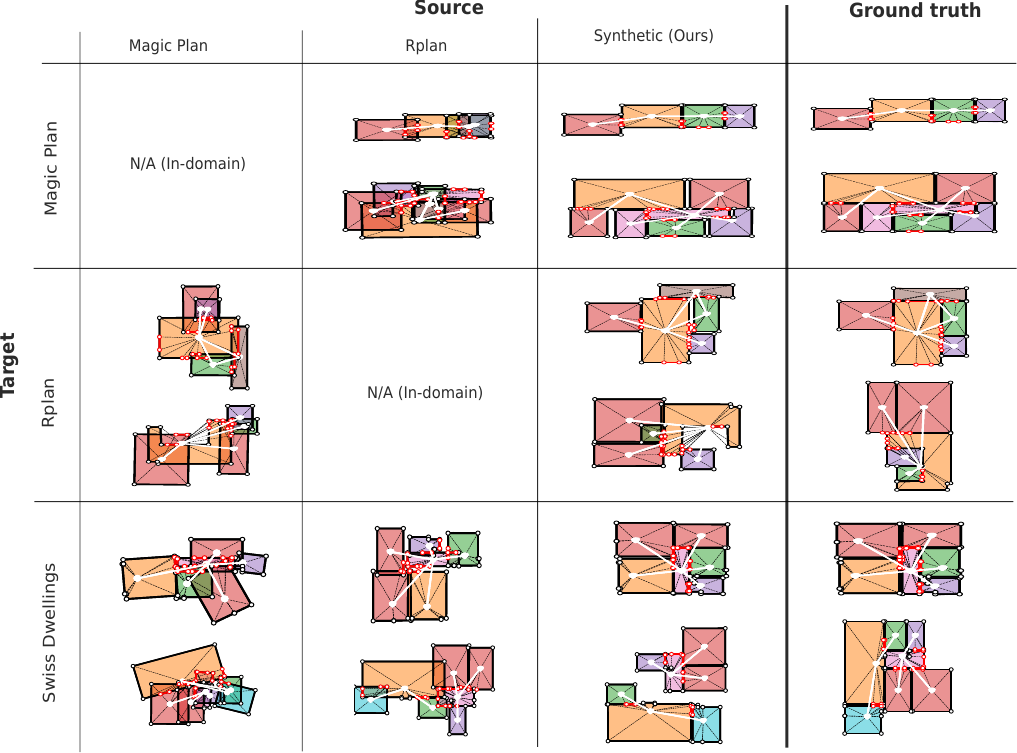}
    \caption{Qualitative zero-shot transfer (DPFM). Rows: target domains; columns: training sources (in-domain omitted). Real-world baselines consistently collapse. For transparency, our synthetic prior pairs a success (top) and failure (bottom) case per domain. Since DPFM strictly preserves rigid geometries, typical failures involve rooms erroneously embedding into one another despite valid connectivity (e.g., RPLAN). On Swiss Dwellings, missing inter-component constraints prevent the global positioning of disconnected sub-graphs.}
    \label{fig:qualitative_dpfm}
\end{figure}

\begin{figure}[h]
    \centering
    \includegraphics[width=\textwidth]{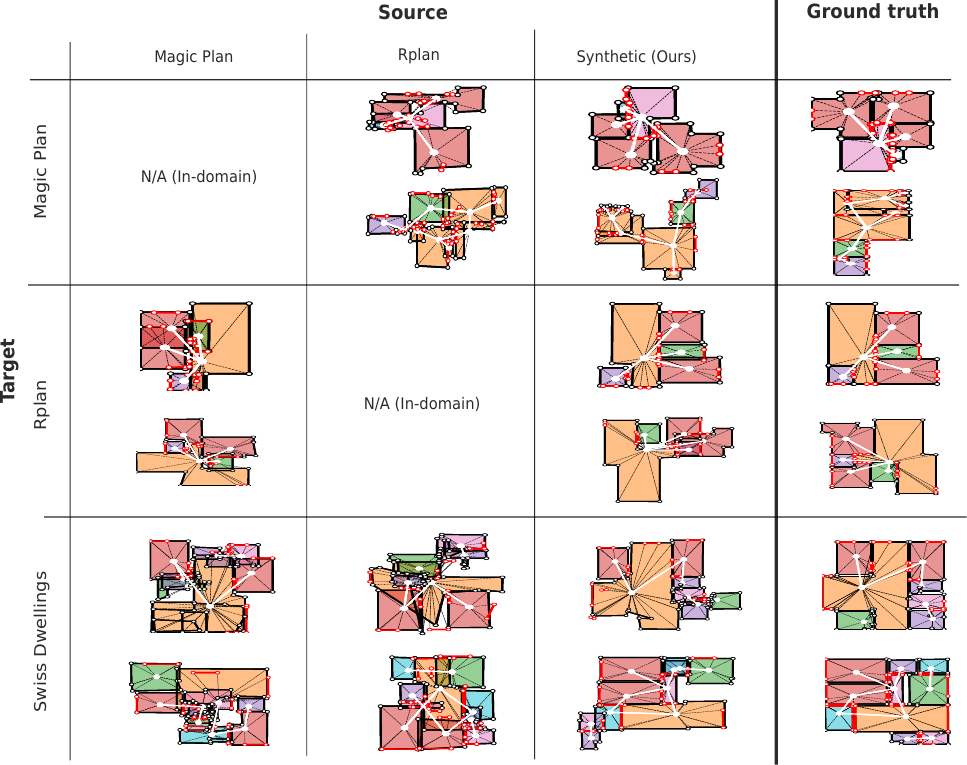}
    \caption{Qualitative zero-shot transfer results (Gueze et al.~\cite{gueze2023floor}). Grid layout strictly follows \Cref{fig:qualitative_dpfm}, displaying a success (top) and failure (bottom) case for our synthetic prior. While cross-domain baselines consistently produce degenerate layouts, our pre-training improves relative arrangements despite visible imperfections. Notably, failures on Swiss Dwellings reflect the vertex-level paradigm's vulnerability to room shapes with highly proximate vertices.}
    \label{fig:qualitative_fpge}
\end{figure}

\begin{table}[t]
    \centering
    \small
    \caption{Zero-shot cross-domain transfer performance (DPFM~\cite{dpfm}). We report MPE $\downarrow$ and NGED $\downarrow$. The OOD Avg. column averages performance exclusively on unseen target domains. Our synthetic pre-training significantly lowers the OOD error compared to real-world baselines, demonstrating its effectiveness as a robust initialization.}
    \label{tab:zeroshot_dpfm}
    \begin{tabular}{@{}llcccc@{}}
        \toprule
        \textbf{Prior (Training)} & \textbf{Metric} & \textbf{RPLAN} & \textbf{MagicPlan} & \textbf{Swiss Dw.} & \textbf{OOD Avg.} \\
        \midrule
        \multirow{2}{*}{\textbf{RPLAN}} 
        & MPE $\downarrow$ & $\mathit{6.31}_{\pm 0.11}$ & $49.33_{\pm 0.25}$ & $52.33_{\pm 1.02}$ & $50.83$ \\
        & NGED $\downarrow$ & $\mathit{0.46}_{\pm 0.01}$ & $0.90_{\pm 0.01}$ & $0.73_{\pm 0.0}$ & \\
        \midrule
        \multirow{2}{*}{\textbf{MagicPlan}} 
        & MPE $\downarrow$ & $57.16_{\pm 0.29}$ & $\mathit{29.10}_{\pm 0.45}$ & $59.19_{\pm 0.73}$ & $58.18$ \\
        & NGED $\downarrow$ & $0.80_{\pm 0.01}$ & $\mathit{0.33}_{\pm 0.0}$ & $0.88_{\pm 0.01}$ &  \\
        \midrule
        \multirow{2}{*}{\textbf{Synthetic (Ours)}} 
        & MPE $\downarrow$ & $\mathbf{38.20}_{\pm 0.27}$ & $\mathbf{23.32}_{\pm 0.43}$ & $\mathbf{49.41}_{\pm 1.43}$ & $\mathbf{36.98}$ \\
        & NGED $\downarrow$ & $\mathbf{0.40}_{\pm 0.01}$ & $\mathbf{0.42}_{\pm 0.01}$ & $\mathbf{0.63}_{\pm 0.01}$ &  \\
        \bottomrule
    \end{tabular}
\end{table}

\begin{table}[t]
    \centering
    \small
    \caption{Zero-shot cross-domain transfer performance (Gueze et al.~\cite{gueze2023floor}). We report MPE $\downarrow$ only (see \Cref{sec:zero_shot}). Trends closely mirror those observed with DPFM in \Cref{tab:zeroshot_dpfm}: real-world baselines suffer from severe bidirectional domain shift, whereas our synthetic pre-training substantially mitigates this collapse on RPLAN and MagicPlan; on Swiss Dwellings the RPLAN prior remains slightly better, consistent with the harder, fragmented topology of that domain.}
    \label{tab:zeroshot_gueze}
    \begin{tabular}{@{}lcccc@{}}
        \toprule
        \textbf{Prior (Training)} & \textbf{RPLAN} & \textbf{MagicPlan} & \textbf{Swiss Dw.} & \textbf{OOD Avg.} \\
        \midrule
        \textbf{RPLAN} & $\mathit{4.95}_{\pm 0.50}$ & $63.00_{\pm 0.90}$ & $\mathbf{51.20}_{\pm 1.05}$ & $57.10$ \\
        \textbf{MagicPlan} & $77.90_{\pm 1.23}$ & $\mathit{4.40}_{\pm 0.90}$ & $73.60_{\pm 1.56}$ & $75.75$ \\
        \midrule
        \textbf{Synthetic (Ours)} & $\mathbf{20.20}_{\pm 0.60}$ & $\mathbf{30.90}_{\pm 1.61}$ & $53.80_{\pm 1.55}$ & $\mathbf{34.97}$ \\
        \bottomrule
    \end{tabular}
\end{table}

\paragraph{Quantitative Analysis.} Table~\ref{tab:zeroshot_dpfm} reports zero-shot cross-domain transfer performance 
across all domain pairs for the DPFM model. We first observe that domain shift in conditioned floor plan 
generation is a systematic and bidirectional phenomenon: models trained on any single 
domain consistently fail to generalize to the others, regardless of the direction of 
transfer. RPLAN-trained models suffer a dramatic performance collapse on MagicPlan 
($6.31 \rightarrow 49.33$) and Swiss Dwellings ($6.31 \rightarrow 52.33$), while 
MagicPlan-trained models degrade equally severely on RPLAN ($29.10 \rightarrow 57.16$) 
and Swiss Dwellings ($29.10 \rightarrow 59.19$). 

Our synthetic training strategy significantly mitigates this collapse. Without 
observing any real floor plan data, our approach achieves the best zero-shot 
performance on MagicPlan ($23.32$), outperforming even direct in-domain 
training ($29.10$), a counter-intuitive result that demonstrates the 
quality of the initialization learned from synthetic data. On Swiss Dwellings, 
our approach achieves the best cross-domain result ($49.41$), ahead of both 
RPLAN-trained ($52.33$) and MagicPlan-trained ($59.19$) baselines. 

On RPLAN, synthetic pre-training yields $38.20$, which remains above the 
in-domain baseline ($6.31$). This gap is expected: RPLAN is a large, 
highly curated dataset with strong geometric regularities that in-domain training 
can exploit directly. We show in Section~\ref{sec:finetuning} that fine-tuning from our synthetic dataset closes this gap substantially. As shown in \Cref{tab:zeroshot_gueze}, we observe consistent trends for the vertex-level paradigm~\cite{gueze2023floor}, confirming that this vulnerability is consistent across both paradigms. Note that we exclusively report MPE for this model: under severe shift, the generated door segments fail to respect strict placement constraints relative to the room geometry. Consequently, severely displaced rooms can accidentally intersect with these decoupled segments, yielding false connectivities that completely skew the topological evaluation (NGED). As visually evident in \Cref{fig:qualitative_fpge}, cross-domain baselines frequently generate doors floating far outside valid boundaries.

\paragraph{Qualitative Failure Analysis.} Visualizations in \Cref{fig:qualitative_dpfm} and \Cref{fig:qualitative_fpge} highlight the distinct failure modes under domain shift. While real-world baselines consistently produce degenerate layouts, our synthetic prior yields coherent arrangements but exhibits specific failure patterns. For DPFM, which strictly preserves rigid input geometries, typical failures consist of rooms inappropriately embedding into one another despite maintaining valid graph connectivity (e.g., on RPLAN). Conversely, performance on Swiss Dwellings is globally lower across all models due to its fragmented topology and complex geometries. Without inter-component spatial constraints, DPFM struggles to globally position disconnected sub-graphs. Furthermore, this dataset features intricate room shapes with highly proximate vertices. This high-frequency geometry severely disrupts vertex-level models like Gueze et al., whose sequence-based representations (chain codes) fail to resolve micro-segments under domain shift. These results confirm that models inherently overfit the architectural details of their training domain rather than learning universal assembly rules; our synthetic pre-training substantially mitigates this bias and reduces cross-domain degradation.

\subsection{Data-Efficient Fine-Tuning}
\label{sec:finetuning}

\begin{figure}[h!]
    \centering
    
    \begin{subfigure}{\textwidth}
        \centering
   
        \includegraphics[width=\textwidth]{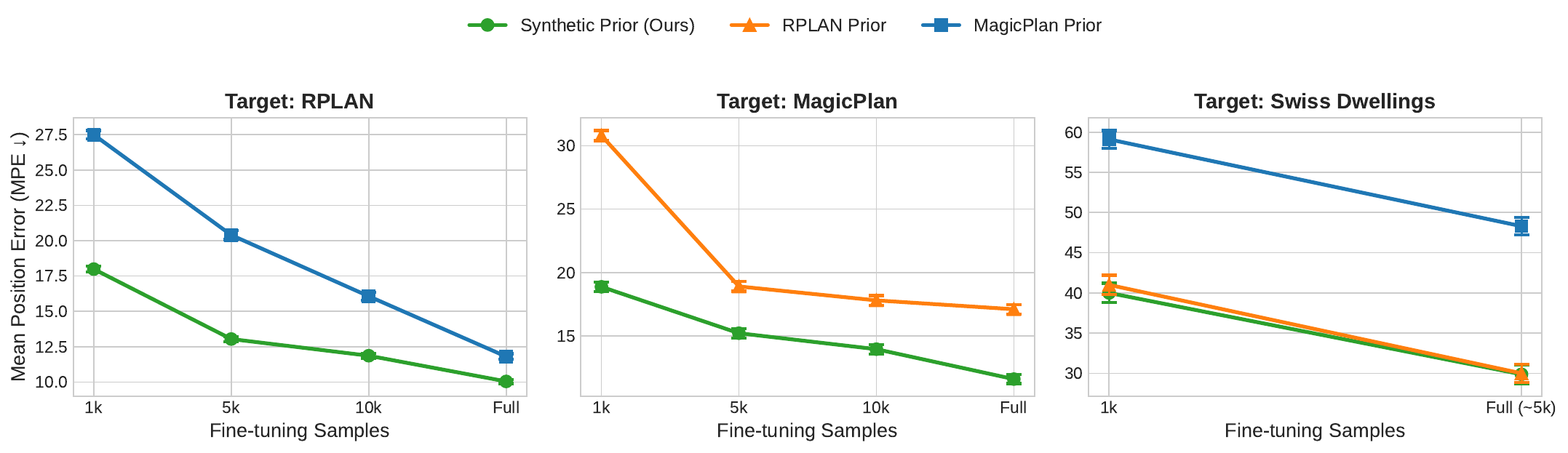}
        \caption{Geometric transfer: Mean Position Error (MPE $\downarrow$)}
        \label{fig:ft_mpe}
    \end{subfigure}

    \begin{subfigure}{\textwidth}
        \centering

        \includegraphics[width=\textwidth]{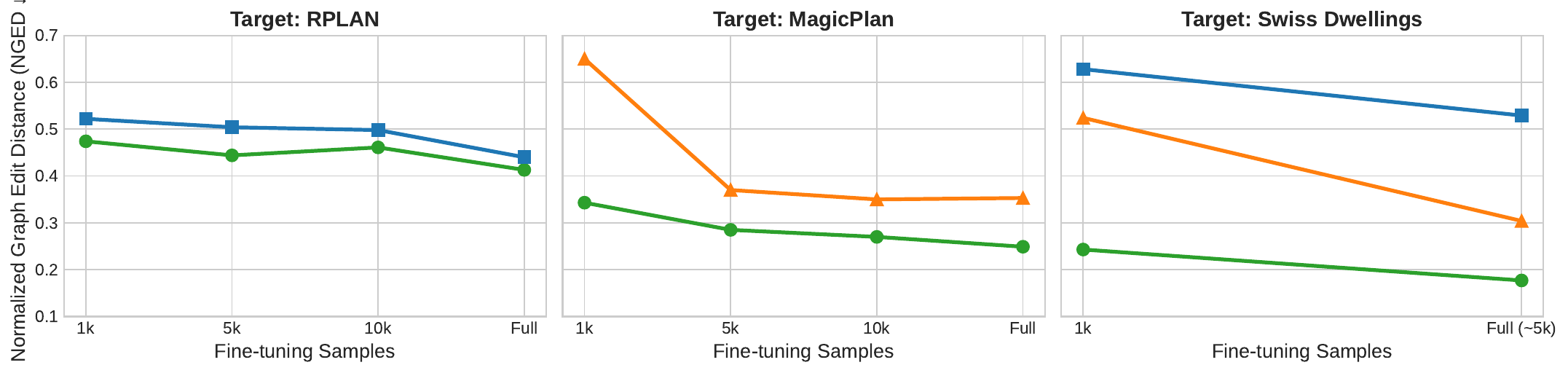}
        \caption{Topological transfer: Normalized Graph Edit Distance (NGED $\downarrow$)}
        \label{fig:ft_nged}
    \end{subfigure}

    \caption{\textbf{Data-efficient fine-tuning and domain adaptation scaling (DPFM).} 
    We compare models initialized with different priors across varying amounts of target data. 
    \textbf{(a)} Geometric performance. \textbf{(b)} Topological performance.
    Across both metrics, our style-agnostic Synthetic prior (green) mitigates negative transfer and outperforms real-world priors (RPLAN and MagicPlan) in low-data regimes, remaining competitive when fine-tuned on the full datasets.}
    \label{fig:finetuning_complete}
\end{figure}

\begin{table}[t]
    \centering
    \small
    \caption{Fine-tuning data efficiency (Gueze et al.~\cite{gueze2023floor}, MPE $\downarrow$). Synthetic pre-training consistently outperforms cross-domain real initialization, confirming that data-efficiency benefits hold across both paradigms. We restrict the Gueze et al. fine-tuning study to RPLAN and MagicPlan up to $10$k samples, where the trend is already established; Swiss Dwellings is omitted because its highly proximate vertices and micro-segments are poorly resolved by the chain-code representation (Sec.~\ref{sec:zero_shot}).}
    \label{tab:ft_gueze}
    \begin{tabular}{@{}lcccccc@{}}
        \toprule
        & \multicolumn{3}{c}{\textbf{Target: RPLAN}} & \multicolumn{3}{c}{\textbf{Target: MagicPlan}} \\
        \cmidrule(lr){2-4} \cmidrule(lr){5-7}
        \textbf{Pre-training} & \textbf{1k} & \textbf{5k} & \textbf{10k} & \textbf{1k} & \textbf{5k} & \textbf{10k} \\
        \midrule
        \textbf{RPLAN} & -- & -- & -- & $19.7_{\pm 1.70}$ & $14.5_{\pm 1.59}$ & $14.1_{\pm 1.46}$ \\
        \textbf{MagicPlan} & $12.1_{\pm 1.24}$ & $7.04_{\pm 1.08}$ & $6.84_{\pm 0.97}$ & -- & -- & -- \\
        \midrule
        \textbf{Synthetic (Ours)} & $\mathbf{7.05}_{\pm 0.90}$ & $\mathbf{6.57}_{\pm 1.09}$ & $\mathbf{5.38}_{\pm 0.80}$ & $\mathbf{13.7}_{\pm 1.40}$ & $\mathbf{12.4}_{\pm 1.73}$ & $\mathbf{12.3}_{\pm 1.40}$ \\
        \bottomrule
    \end{tabular}
\end{table}


While synthetic zero-shot transfer significantly mitigates domain shift, a gap with in-domain training remains. To determine if minimal target data can close this gap, we fine-tune both models on subsets of size $N \in \{1\text{k}, 5\text{k}, 10\text{k}, |\mathcal{D}^t|\}$ (restricted to $N \in \{1\text{k}, |\mathcal{D}^t|\}$ for the smaller Swiss Dwellings), comparing all initializations under identical conditions.

\paragraph{Geometric Transfer.} As illustrated in \Cref{fig:finetuning_complete}a, our synthetic initialization yields the lowest error across fine-tuning regimes. The performance gap is particularly wide in low-data settings: at 1k samples, fine-tuning from RPLAN to MagicPlan results in an MPE of 30.8, whereas our synthetic prior reaches 18.8. A similar trend is observed for MagicPlan-to-RPLAN transfer, where our prior achieves 17.9 MPE at 1k samples compared to 27.5 for the real-world baseline. On the more complex Swiss Dwellings dataset, the performance of the RPLAN and synthetic priors converges during fine-tuning. We attribute this to a shared limitation: neither RPLAN nor our synthetic dataset contains the highly proximate vertices and micro-segments characteristic of Swiss Dwellings. This lack of high-frequency geometric detail in pre-training likely creates a common performance ceiling. While our synthetic prior maintains a superior zero-shot starting point, target-domain supervision allows the RPLAN prior to resolve its global positioning issues, leading both models to encounter the same limits of geometric precision. In contrast, the MagicPlan prior fails to close this gap, suggesting that its internal representations are too constrained by its original domain to adapt to such complex topological structures.

\paragraph{Generalization across paradigms.} To verify that this data-efficiency is consistent across paradigms, we evaluate the vertex-level diffusion paradigm~\cite{gueze2023floor}. As reported in \Cref{tab:ft_gueze}, synthetic pre-training consistently improves fine-tuning performance, achieving significantly lower MPE than real-world cross-domain initializations on both RPLAN and MagicPlan. Notably, on RPLAN with only 1k samples, our prior yields an MPE of 7.05 compared to 12.1 for the MagicPlan-initialized model, a 41.7\% reduction in error. These results confirm that the robust spatial priors learned from synthetic data provide a transferable initialization that accelerates adaptation across different generative architectures.

\paragraph{Topological Consistency.} This geometric advantage is also reflected in the topological correctness (NGED) for the DPFM model. As shown in \Cref{fig:finetuning_complete}b, the synthetic initialization maintains the lowest NGED across nearly all data splits. Interestingly, on Swiss Dwellings, the synthetic prior achieves a better NGED than the RPLAN baseline, even when their MPE scores converge. Since NGED measures the connectivity derived from the intersection of generated door segments and room boundaries, this gap suggests that our prior yields a more precise local alignment between doors and geometry. This improved consistency confirms that the assembly rules learned from synthetic data transfer effectively, even in complex domains where global geometric adaptation reaches a plateau.

Overall, our synthetic data provides a superior starting point by providing models with the fundamental rules of floor plan assembly. This lead in data efficiency means that fewer real-world samples are required to reach competitive performance, a benefit that holds across different generative paradigms.

\section{Discussion and Limitations}

\begin{table}[t]
\centering
\small
\caption{Ablation on synthetic dataset creation pipeline (DPFM, zero-shot MPE $\downarrow$). We evaluate the impact of dataset size and shape complexity. Removing shape augmentations (simple shapes) dramatically degrades transfer. Increasing dataset size from 60k to 135k yields diminishing returns.}
\label{tab:ablation}
\begin{tabular}{lcccc}
\toprule
Configuration & RPLAN & MagicPlan & Swiss Dw. & Avg. \\
\midrule
135k, complex shapes & $38.20_{\pm 0.27}$ & $\mathbf{23.32}_{\pm 0.43}$ & $\mathbf{49.41}_{\pm 1.43}$ & $36.98$ \\
60k, complex shapes & $\mathbf{35.80}_{\pm 0.24}$ & $24.70_{\pm 0.42}$ & $49.50_{\pm 1.30}$ & $\mathbf{36.67}$ \\
60k, simple shapes & $69.03_{\pm 0.31}$ & $43.70_{\pm 0.44}$ & $67.80_{\pm 1.06}$ & $60.18$ \\
\bottomrule
\end{tabular}
\end{table}

While our synthetic pre-training significantly improves cross-domain transfer, a gap remains with in-domain training, particularly on RPLAN ($6.31$ vs $38.20$ MPE). This is expected given RPLAN's high geometric regularity, which in-domain models can easily capture. Future work could explore more advanced pre-training strategies, such as adaptive augmentation schemes targeting specific failure modes, to further narrow this gap.

Our ablation study (\Cref{tab:ablation}) shows that shape-level diversity is the most critical factor for generalization. Removing shape augmentations (simple shapes) leads to a massive performance drop, confirming that "unrealistic" geometric complexity is necessary to learn robust assembly rules. Interestingly, increasing the dataset size from 60k to 135k yields diminishing returns, suggesting that the variety of the assembly strategy is more important than raw data volume. The high corner count of our synthetic rooms (\Cref{tab:dataset_stats}) is a deliberate design choice, not a defect: real-domain regularities saturate in-domain (RPLAN $6.31$ MPE) but fail OOD, whereas geometric implausibility yields transferable assembly rules; removing it costs $+23$ average MPE (\Cref{tab:ablation}).

A remaining challenge concerns datasets with fragmented connectivity graphs or high-frequency details, such as Swiss Dwellings. First, we observe that models successfully assemble individual connected components but struggle to position disconnected sub-graphs relative to each other. This is a limitation of the conditioning signal: without edges between components, no spatial relationship is defined in the input graph. Incorporating global constraints, such as bounding boxes, could provide the missing context. Second, the presence of extremely proximate vertices in such datasets remains difficult for both the rigid transformation (DPFM) and vertex-level (Gueze et al.) paradigms. Since our synthetic pipeline and most real-world datasets favor cleaner geometries, these configurations are simply not encountered during training, leading to lower precision.

\begin{table}[t]
\centering\small
\caption{Shape-pool independence (zero-shot MPE $\downarrow$). We re-seed the \emph{identical} assembly pipeline with strictly orthogonal ProcTHOR~\cite{procthor} room shapes instead of RPLAN polygons. Gueze et al., whose chain-code angle regularization is pool-robust, transfers almost identically across seeds. DPFM lacks this regularization: a small-rotation augmentation ($\dagger$) recovers RPLAN/MagicPlan transfer, while the Swiss Dwellings residual reflects pool properties beyond global obliquity and remains a limitation of this ablation.}
\label{tab:shape_pool}
\begin{tabular}{@{}llccc@{}}
\toprule
\textbf{Model} & \textbf{Seed pool} & \textbf{RPLAN} & \textbf{MagicPlan} & \textbf{Swiss Dw.} \\
\midrule
\multirow{2}{*}{Gueze et al.~\cite{gueze2023floor}} & RPLAN    & $20.2$ & $30.9$ & $53.8$ \\
                                                    & ProcTHOR & $23.7$ & $31.2$ & $53.7$ \\
\midrule
\multirow{2}{*}{DPFM~\cite{dpfm}} & RPLAN               & $38.2$ & $23.3$ & $49.4$ \\
                                  & ProcTHOR$^{\dagger}$ & $37.4$ & $27.6$ & $63.0$ \\
\bottomrule
\end{tabular}
\end{table}

A natural concern is that gains stem from RPLAN-specific shape priors, since the synthetic pool is seeded from isolated RPLAN polygons (Sec.~\ref{sec:pre_train_dataset}). We re-seed the same pipeline with ProcTHOR~\cite{procthor} shapes, which are strictly orthogonal, in contrast to the slight non-orthogonality of RPLAN that is also present in all three targets. As shown in \Cref{tab:shape_pool}, Gueze et al.\ transfers almost identically under both seeds (e.g., $53.8\!\to\!53.7$ on Swiss Dwellings), indicating that the benefit is carried by the assembly pipeline, not the source pool. DPFM, which lacks chain-code regularization, recovers RPLAN and MagicPlan transfer with a small-rotation augmentation, but its Swiss Dwellings residual grows, a limitation of the current ablation. The cross-seed robustness of the constraint-based paradigm supports our central claim that transferability arises from spatial-assembly rules rather than RPLAN-specific priors.

\section*{Conclusion}

We presented the first study of domain shift in conditioned floor plan generation, showing that models from two paradigms suffer severe performance degradation when transferred across datasets. To address this, we introduced a procedural synthetic pipeline that enforces physical constraints while sacrificing architectural realism through geometric augmentation of layout and room shapes.

Our experiments across three public datasets show that pre-training on this synthetic data yields a robust initialization: in zero-shot transfer, it consistently outperforms real cross-domain pre-training and even surpasses in-domain training on MagicPlan with DPFM. Crucially, our synthetic initialization significantly accelerates adaptation in low-data regimes: when fine-tuned with as few as 1k to 10k samples, it consistently outperforms real-world initializations using the same amount of data. This drastically reduces the need for large-scale dataset acquisition when deploying models in new architectural domains.

Beyond floor plan generation, we believe our findings highlight a broader principle for constrained geometric generation tasks: learning from diverse, rule-compliant but visually implausible configurations can produce more transferable representations than training on realistic but domain-specific data. Extending this approach to other layout domains, such as molecules arrangement or puzzle solving, and investigating how to close the remaining gap with in-domain training without target-domain supervision, are promising directions for future work.

\section*{Acknowledgments}

This project was provided with computing HPC and storage resources by GENCI at IDRIS thanks to the grant 2026-AD011016120R1 on the supercomputer Jean Zay.

\bibliographystyle{plain}
\bibliography{ref_eccv}
\end{document}